\documentclass[11pt]{article}

\usepackage[preprint]{acl}

\usepackage{times}
\usepackage{latexsym}

\usepackage[T1]{fontenc}

\usepackage[utf8]{inputenc}

\usepackage{microtype}

\usepackage{inconsolata}

\usepackage{graphicx}

\usepackage{tcolorbox}
\usepackage{url}            
\usepackage{booktabs}       
\usepackage{amsfonts}       
\usepackage{nicefrac}       
\usepackage{xcolor}         
\tcbuselibrary{breakable}
\usepackage{dashrule}
\tcbuselibrary{skins}

\usepackage{amsmath}
\usepackage{amssymb}
\usepackage{algorithm}
\usepackage{algorithmic}
\usepackage{multirow}
\usepackage{subcaption}
\usepackage{enumitem}
\usepackage{pifont}
\usepackage{colortbl}
\usepackage{soul}
\usepackage{mdframed}
\usepackage{caption}
\usepackage{hyperref}       

\title{Evaluate-as-Action: Self-Evaluated Process Rewards for Retrieval-Augmented Agents}

\author{%
	Jiangming Shu\textsuperscript{1}, 
	Yuxiang Zhang\textsuperscript{1}, 
    Ye Ma\textsuperscript{2}, 
    Xueyuan Lin\textsuperscript{2},
	Jitao Sang\textsuperscript{1}\thanks{Corresponding author.} \\
	\textsuperscript{1}School of Computer Science and Technology, Beijing Jiaotong University \\
	\textsuperscript{2}Hithink Research \\
	\texttt{\{jiangmingshu, yuxiangzhang, jtsang\}@bjtu.edu.cn} \\
    \texttt{maye@myhexin.com}\\
    \texttt{linxy59@mail2.sysu.edu.cn}
}

\begin{document}
\maketitle
\begin{abstract}
Retrieval-augmented agents can query external evidence, yet their reliability in multi-step reasoning remains limited: noisy retrieval may derail multi-hop question answering, while outcome-only reinforcement learning provides credit signals that are too coarse to optimize intermediate steps.
We propose \textsc{EvalAct} (Evaluate-as-Action), which converts implicit retrieval quality assessment into an explicit action and enforces a coupled Search-to-Evaluate protocol so that each retrieval is immediately followed by a structured evaluation score, yielding process signals aligned with the interaction trajectory.
To leverage these signals, we introduce Process-Calibrated Advantage Rescaling (PCAR), a GRPO-based optimization method that rescales advantages at the segment level according to evaluation scores, emphasizing reliable segments while updating uncertain ones conservatively.
Experiments on seven open-domain QA benchmarks show that \textsc{EvalAct} achieves the best average accuracy, with the largest gains on multi-hop tasks, and ablations verify that the explicit evaluation loop drives the primary improvements while PCAR provides consistent additional benefits. 
\end{abstract}

\section{Introduction}

Large language model (LLM) agents have shifted automated reasoning from passive response generation to autonomous problem solving, where models plan, interact with external tools, and iteratively refine their beliefs across multi-step trajectories~\citep{react,toolformer}.
Retrieval-augmented generation (RAG) further extends this capability by grounding decisions in external evidence, enabling open-domain question answering beyond the limits of parametric knowledge~\citep{lewis2020retrieval,guu2020retrieval}.
However, as queries shift from single-hop factoids to multi-hop narratives, the central bottleneck is no longer tool access itself, but the agent's ability to navigate, verify, and synthesize evidence over long-horizon, noise-prone interaction sequences~\citep{ircot,selfrag}.

Despite substantial progress, ensuring reliable intermediate reasoning remains a key challenge.
Existing agentic baselines, from prompting methods that interleave retrieval and reasoning~\citep{ircot} to RL-based search agents such as Search-R1~\citep{searchr1} and refinement frameworks such as AutoRefine~\citep{autorefine}, still rely primarily on implicit internal reasoning for noise suppression and self-correction.
This paradigm suffers from two fundamental limitations.
First, \textbf{error propagation}: without an explicit, immediate mechanism for evidence verification, a single irrelevant document can derail downstream reasoning, causing irreversible trajectory drift in multi-hop settings.
Second, \textbf{coarse credit assignment}: standard RL optimization, including PPO-based RLHF~\citep{ouyang2022training,ppo} and outcome-reward post-training methods such as GRPO~\citep{deepseekmath}, typically relies on sparse signals tied to final-answer correctness.
Such outcome-only supervision cannot distinguish informative retrieval steps from redundant or misleading actions within long trajectories; as a result, the optimizer often reinforces or penalizes an entire trajectory nearly uniformly, degrading sample efficiency and causing performance saturation as task complexity grows.

To address these challenges, we introduce \textsc{EvalAct}, a reinforcement learning framework that transforms the agent's implicit self-assessment of retrieval quality into an explicit, policy-selectable action.
\textsc{EvalAct} enforces a strictly coupled search-then-evaluate protocol: each \texttt{Search} action must be immediately followed by an \texttt{Evaluate} action that produces a structured self-assessment score reflecting the utility of the retrieved evidence.
This design directly addresses the two limitations identified above. 
At inference time, the evaluation output provides actionable control signals that facilitate early pruning of unproductive branches, reducing error propagation without external oracle supervision. 
During training, it produces dense, trajectory-aligned process signals that make intermediate reliability directly optimizable and enable finer-grained credit assignment.

To leverage these process signals effectively, we further propose \textbf{Process-Calibrated Advantage Rescaling (PCAR)} built upon Group Relative Policy Optimization (GRPO)~\citep{deepseekmath}.
Instead of broadcasting a single trajectory-level advantage to all tokens, PCAR uses step-wise self-evaluation scores to modulate updates at the segment level, amplifying gradients for reliable, progress-making steps while applying conservative updates to uncertain segments.
Importantly, this provides process-level guidance without requiring expensive human-annotated process reward models~\citep{lightman2023let}, while complementing prior verification-oriented supervision that does not explicitly target retrieval behavior~\citep{ma2502s2r}.
Together, \textsc{EvalAct} and PCAR convert introspection into an executable action space with trainable process signals, improving learning stability and multi-hop generalization.

Our contributions are as follows:
\begin{itemize}[leftmargin=*]
    \item We propose \textsc{EvalAct}, an RL framework that transforms implicit retrieval quality evaluation into an explicit \texttt{Evaluate} action and enforces a coupled \texttt{Search}$\rightarrow$\texttt{Evaluate} protocol, producing dense, trajectory-aligned self-evaluation rewards for tool-using agents.
    \item We introduce Process-Calibrated Advantage Rescaling (PCAR), a GRPO-based optimization strategy that leverages step-wise self-evaluation scores to refine credit assignment and stabilize learning in long-horizon retrieval trajectories.
    \item We achieve the best average performance across seven open-domain QA benchmarks with two backbone scales, with particularly strong gains on multi-hop tasks; extensive ablations show that the explicit evaluation loop accounts for the dominant improvements, while PCAR provides consistent additional benefits.
\end{itemize}

\section{Methodology}
\label{sec:method}

\begin{figure*}[!t]
  \centering
  \includegraphics[width=\linewidth]{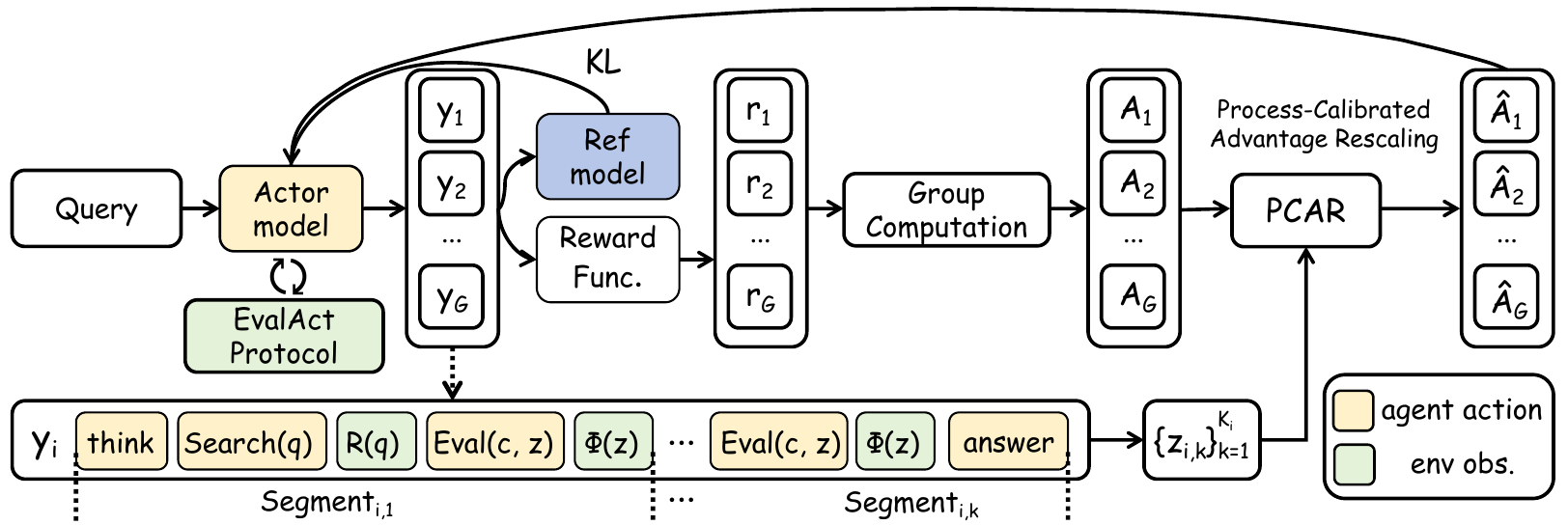}
    \caption{\textbf{Overview of \textsc{EvalAct} with PCAR.}
The agent follows a coupled \texttt{Search}$\rightarrow$\texttt{Evaluate} protocol, producing segment-wise self-evaluation scores $\{z_{i,k}\}$ that PCAR uses to rescale GRPO advantages.}
  \label{fig:method}
\end{figure*}

We present our approach in three parts.
First, we formulate retrieval-augmented multi-hop question answering as sequential decision-making under partial observability, providing a unified view for both inference-time interaction and RL training (\S\ref{ssec:formulation}).
Second, we introduce \textsc{EvalAct} (Evaluate-as-Action), which transforms implicit retrieval quality evaluation into an explicit, policy-selectable action and enforces a coupled \texttt{Search}$\rightarrow$\texttt{Evaluate} interaction protocol (\S\ref{ssec:framework}).
Third, we propose Process-Calibrated Advantage Rescaling (PCAR), a GRPO-based optimization method that rescales segment-wise policy gradients using self-evaluation scores, improving credit assignment and stabilizing learning (\S\ref{ssec:optimization}). Figure~\ref{fig:method} illustrates the coupled \texttt{Search}$\rightarrow$\texttt{Evaluate} loop and the PCAR-weighted GRPO update.

\subsection{Problem Formulation}
\label{ssec:formulation}

Let $\mathcal{M}_{\theta}$ be an LLM parameterized by $\theta$, inducing a stochastic policy $\pi_{\theta}$ over textual tokens and tool-mediated actions.
We model retrieval-augmented multi-hop question answering as a POMDP $\langle \mathcal{S}, \mathcal{A}, \mathcal{O}, \mathcal{P} \rangle$.
At time $t=0,\ldots,T$, the agent samples an action $a_t \sim \pi_{\theta}(\cdot \mid h_t)$ conditioned on the observable history
\begin{equation}
    h_t = [\,x, a_0, o_0, \ldots, a_{t-1}, o_{t-1}\,],
\end{equation}
where $x$ is the input query and $o_t \in \mathcal{O}$ is the observation returned by the environment after executing $a_t$.
A trajectory is $\tau=\{(a_t,o_t)\}_{t=0}^{T}$, while the underlying state $s_t \in \mathcal{S}$ is unobserved.
The transition function $\mathcal{P}(s_{t+1} \mid s_t, a_t)$ governs state evolution and is implicitly defined by the retrieval environment and the agent's reasoning process.

\paragraph{Action space.}
We partition $\mathcal{A}$ into (i) reasoning tokens $\mathcal{A}_{\text{think}}$, (ii) tool actions $\mathcal{A}_{\text{tool}}$, and (iii) a terminal answer action $\mathcal{A}_{\text{answer}}$.
Tool actions include retrieval $\texttt{Search}(q)$ and self-evaluation $\texttt{Evaluate}(c,z)$, where $q$ is a query string, $c$ is a textual assessment, and $z \in [0,10]$ is a scalar confidence score reported by the policy.

\paragraph{Observations.}
For tool actions, the environment returns
\begin{equation}
    o_t \sim 
    \begin{cases}
        \mathcal{R}(q), & \text{if } a_t = \texttt{Search}(q), \\[2pt]
        \Phi(z), & \text{if } a_t = \texttt{Evaluate}(c,z), \\[2pt]
        \emptyset, & \text{otherwise},
    \end{cases}
\end{equation}
where $\mathcal{R}(q)$ denotes the top-$k$ retrieved documents and $\Phi(\cdot)$ maps the reported score to a discrete feedback cue used for subsequent decision-making.

\subsection{\textsc{EvalAct}: Evaluate-as-Action}
\label{ssec:framework}

\textsc{EvalAct} transforms implicit retrieval quality evaluation into an executable action and couples each retrieval step with immediate self-assessment.
Specifically, after any retrieval action $a_t=\texttt{Search}(q)$ with observation $o_t=\mathcal{R}(q)$, the agent must invoke exactly one evaluation action $a_{t+1}=\texttt{Evaluate}(c,z)$.
The assessment $c$ is conditioned on the retrieved documents, and $z\in[0,10]$ is a self-reported confidence score.
This one-to-one coupling aligns each retrieval result with an explicit reliability assessment, enabling segment-wise training signals.

\paragraph{Inference-time control without oracle signals.}
To avoid external supervision, the environment-side evaluator is deliberately non-interpretive: it neither parses $c$ nor inspects retrieved documents.
Instead, it deterministically maps $z$ to a discrete control cue $\mathcal{I}=\Phi(z)$:
\begin{equation}
    \Phi(z) =
    \begin{cases}
        \mathcal{I}_{\text{low}}, & z \in [0,3], \\[2pt]
        \mathcal{I}_{\text{mid}}, & z \in (3,7], \\[2pt]
        \mathcal{I}_{\text{high}}, & z \in (7,10].
    \end{cases}
\end{equation}
The cue $\mathcal{I}$ is appended to the context and modulates subsequent actions via instruction conditioning.
For completeness, Appendix~\ref{app:evaluate_templates} specifies the Evaluate format and feedback templates, and Appendix~\ref{app:case_study} presents a complete multi-hop trajectory showing how the agent assigns calibrated self-evaluation scores to guide iterative evidence gathering.

\subsection{Reinforcement Learning with PCAR}
\label{ssec:optimization}

We optimize $\pi_{\theta}$ to maximize expected reward, using GRPO as the backbone and PCAR to incorporate process signals from $\texttt{Evaluate}$.

\paragraph{Gated outcome reward.}
To enforce protocol compliance while optimizing answer quality, we use a gated reward:
\begin{equation}
\label{eq:reward}
    \mathcal{G}(y, y^*) = 
    \begin{cases}
        \text{F1}(y_{\text{ans}}, y^*), & \text{if } \mathbb{I}_{\text{fmt}}(y)=1, \\
        0, & \text{otherwise},
    \end{cases}
\end{equation}
where $y_{\text{ans}}$ is the final answer extracted from \texttt{<answer>} tags and $\mathbb{I}_{\text{fmt}}(y)$ indicates whether (i) reasoning is enclosed by \texttt{<think>} tags and (ii) every \texttt{Search} is immediately followed by \texttt{Evaluate}.

\paragraph{GRPO.}
Given $G$ rollouts $\{y_1,\ldots,y_G\}$ sampled from $\pi_{\theta_{\text{old}}}$ for the same input $x$, we compute group-normalized advantages
\begin{equation}
    A_i = \frac{r_i - \mu_{\text{group}}}{\sigma_{\text{group}} + \varepsilon},
\end{equation}
where $r_i=\mathcal{G}(y_i,y^*)$ and $\mu_{\text{group}},\sigma_{\text{group}}$ are the within-group mean and standard deviation.

\paragraph{PCAR: segment-wise advantage rescaling.}
Standard GRPO applies the same $A_i$ to all tokens, which can inadvertently reinforce unreliable intermediate steps.
PCAR instead rescales advantages at the segment level using the self-evaluation scores.
Let $y_i$ contain $K_i$ segments, each associated with a score $z_{i,k}\in[0,10]$.
We first compute an intra-trajectory standardized reliability signal
\begin{equation}
    \tilde{z}_{i,k} = \frac{z_{i,k} - \mu_i}{\sigma_i + \varepsilon},
\end{equation}
where $\mu_i$ and $\sigma_i$ are the mean and standard deviation of $\{z_{i,k}\}_{k=1}^{K_i}$.
This normalization makes $\tilde z_{i,k}$ reflect relative reliability within the trajectory and suppress trivial constant scoring.

We then define a score-scaled gain
\begin{equation}
    \lambda_{i,k} = \lambda_{\text{base}} + (\lambda_{\text{max}} - \lambda_{\text{base}})\cdot \frac{z_{i,k}}{10},
    \label{eq:pcar_lambda}
\end{equation}
and compute the token-level calibrated advantage for any token $t$ belonging to segment $k$:
\begin{equation}
    \hat{A}_{i,t} = A_i \cdot \mathrm{clamp}\!\left(1 + \lambda_{i,k}\tilde{z}_{i,k},\, \delta,\, \infty\right),
\end{equation}
where $\delta>0$ prevents gradient inversion; we set $\delta=10^{-6}$ in all experiments unless otherwise specified.

Finally, we maximize the GRPO-style clipped objective with the calibrated advantages:
\begin{align}
\notag
&\mathcal{J}(\theta)
= {} \mathbb{E}_{x \sim \mathcal{D}, \{y_i\}_{i=1}^{G} \sim \pi_{\theta_{\text{old}}}}
\Biggl[
\frac{1}{G}\sum_{i=1}^{G}\sum_{t=1}^{L_i}
\Bigl(
\mathcal{L}^{\text{CLIP}}_{i,t} \\
\label{eq:pcar_objective}
& \qquad -\beta\, \mathbb{D}_{\text{KL}}\!\left(
\pi_{\theta}(\cdot \mid h_t)\, \|\, \pi_{\theta_{\text{ref}}}(\cdot \mid h_t)
\right)
\Bigr)
\Biggr],
\end{align}
where $L_i$ is the length of $y_i$ and
\begin{equation}
\label{eq:clip_objective}
\resizebox{\linewidth}{!}{$
\begin{aligned}[b]
\mathcal{L}^{\text{CLIP}}_{i,t}
&=
\min \left(
\rho_{i,t}\hat{A}_{i,t},\,
\mathrm{clip}(\rho_{i,t}, 1-\epsilon, 1+\epsilon)\hat{A}_{i,t}
\right) \\
&\qquad
, \rho_{i,t}
=
\frac{\pi_{\theta}(y_{i,t}\mid h_t)}
{\pi_{\theta_{\text{old}}}(y_{i,t}\mid h_t)}.
\end{aligned}
$}
\end{equation}

By steering updates toward segments that are both outcome-aligned and process-reliable, PCAR improves credit assignment in long-horizon retrieval trajectories.
For reproducibility, Appendix~\ref{sec:appendix_evalact_pcar} provides pseudocode for the complete \textsc{EvalAct} training loop with PCAR, including protocol-compliant rollouts, GRPO advantage estimation, and segment-wise advantage rescaling.

\section{Experiments}
\label{experiments}

\subsection{Experimental Setup}
\label{ssec:exp_setup}

\paragraph{Datasets.}
We evaluate open-domain question answering performance on seven widely-used benchmarks spanning both single-hop and multi-hop settings: Natural Questions (NQ)~\citep{kwiatkowski2019natural}, TriviaQA~\citep{triviaqa}, and PopQA~\citep{popqa} as single-hop datasets, and HotpotQA~\citep{hotpotqa}, 2WikiMultihopQA~\citep{2wikiqa}, MuSiQue~\citep{musique}, and Bamboogle~\citep{bamboogle} as multi-hop datasets that typically require iterative evidence acquisition.
For training, we use the publicly released ASearcherBase35K corpus~\citep{asearcher} and remove invalid or non-actionable samples via lightweight filtering, resulting in $27\text{k}$ instances for RL. For supervised warm-up, we synthesize $2\text{k}$ protocol-compliant trajectories by prompting DeepSeek-V3.2 (Non-thinking Mode)~\citep{liu2025deepseek} to follow the \textsc{EvalAct} interaction format.

\paragraph{Baselines.}
We compare \textsc{EvalAct} against representative baselines spanning direct answering, single-pass retrieval augmentation, and multi-step retrieval–reasoning.
(1) \textbf{Direct Generation} uses the instruction-tuned backbone model to answer using only parametric knowledge, without any retrieval.
(2) \textbf{Na\"ive RAG} retrieves documents once and concatenates them with the query, then generates the answer in a single forward pass.
(3) \textbf{IRCoT}~\citep{ircot} interleaves retrieval and chain-of-thought prompting for multi-hop reasoning.
(4) \textbf{Search-o1}~\citep{searcho1} and (5) \textbf{Search-R1}~\citep{searchr1} represent recent search-augmented agentic baselines with iterative retrieval and reasoning.
(6) \textbf{AutoRefine}~\citep{autorefine} is an iterative refinement baseline that alternates between evidence gathering and answer refinement.
For all retrieval-enabled baselines, we use the same retrieval environment as \textsc{EvalAct}, including the corpus, retriever, returned top-$k$ documents, and search budget, ensuring controlled comparison under matched external evidence access.

\paragraph{Evaluation Metrics.}
We report Exact Match (EM) as the primary evaluation metric on all benchmarks, computed via exact string matching between the normalized prediction and the reference answer.
During RL training, the outcome-level reward is defined as the token-level F1 score between the generated answer and the ground-truth reference (cf.~Eq.~\ref{eq:reward}). 
At test time, performance is evaluated using EM to align with standard open-domain QA evaluation protocols.

\subsection{Implementation Details}
We conduct experiments with two backbones, Qwen2.5-3B-Instruct and Qwen2.5-7B-Instruct. Unless otherwise specified, training uses $8$ NVIDIA A100 GPUs with full-parameter optimization and gradient checkpointing.
We use a fixed open-domain retrieval environment built on the December 2018 Wikipedia dump with a standard BM25 retriever, without reranking or post-retrieval filtering.
At each retrieval step, the top-$k=3$ documents are returned and appended to the dialogue context.
The tool budget is capped at $20$ \texttt{Search} calls per question.

\begin{table*}[t]
\centering
\resizebox{\textwidth}{!}{%
\begin{tabular}{lcccccccc}
\toprule[1.5pt]
& \multicolumn{3}{c}{\textbf{Single-Hop QA}} & \multicolumn{4}{c}{\textbf{Multi-Hop QA}} & \textbf{Avg.} \\
\cmidrule(lr){2-4}\cmidrule(lr){5-8}
\textbf{Method} & \textbf{NQ} & \textbf{PopQA} & \textbf{TriviaQA} & \textbf{2Wiki} & \textbf{Bamboogle} & \textbf{HotpotQA} & \textbf{MuSiQue} & \\
\midrule
\multicolumn{9}{l}{\textbf{Backbone: Qwen2.5-3B-Instruct}} \\
\midrule
Direct Generation & 10.6 & 10.8 & 28.8 & 24.4 & 2.4  & 14.9 & 2.0  & 13.4 \\
IRCoT            & 11.1 & 20.0 & 31.2 & 17.1 & 24.0 & 16.4 & 6.7  & 18.1 \\
Search-o1        & 23.8 & 26.2 & 47.2 & 21.8 & 32.0 & 22.1 & 5.4  & 25.5 \\
Na\"ive RAG      & 34.8 & 38.7 & 54.4 & 22.6 & 8.0  & 25.5 & 4.7  & 27.0 \\
Search-R1        & 34.1 & 37.8 & 54.5 & 31.9 & 26.4 & 32.4 & 10.3 & 32.5 \\
AutoRefine       & \textbf{46.7} & \textbf{45.0} & \underline{62.0} & \underline{39.4} & \underline{34.4} & \underline{40.5} & \underline{15.7} & \underline{40.5} \\
EvalAct-3B(ours) & \underline{38.5} & \underline{43.2} & \textbf{62.1} & \textbf{50.0} & \textbf{48.0} & \textbf{44.3} & \textbf{21.6} & \textbf{44.0} \\
\midrule
\multicolumn{9}{l}{\textbf{Backbone: Qwen2.5-7B-Instruct}} \\
\midrule
Direct Generation & 13.4 & 14.0 & 40.8 & 25.0 & 12.0 & 18.3 & 3.1  & 18.1 \\
IRCoT            & 22.4 & 30.1 & 47.8 & 14.9 & 22.4 & 13.3 & 7.2  & 22.6 \\
Search-o1        & 15.1 & 13.1 & 44.3 & 17.6 & 29.6 & 18.7 & 5.8  & 20.6 \\
Na\"ive RAG      & 34.9 & 39.2 & 58.5 & 23.5 & 20.8 & 29.9 & 5.8  & 30.4 \\
Search-R1        & \underline{39.3} & 39.7 & 61.0 & \underline{41.4} & 36.8 & 37.0 & 14.6 & 38.5 \\
AutoRefine       & \textbf{48.4} & \textbf{48.7} & \textbf{65.9} & 40.5 & \underline{51.2} & \underline{45.1} & \underline{18.7} & \underline{45.5} \\
EvalAct-7B(ours) & 38.5 & \underline{43.6} & \underline{65.6} & \textbf{52.1} & \textbf{56.0} & \textbf{48.8} & \textbf{25.3} & \textbf{47.1} \\
\bottomrule
\end{tabular}
}
\caption{\textbf{Main results (EM, \%) on seven open-domain QA benchmarks.} \textbf{Bold} and \underline{underlined} values indicate the best and second-best performance, respectively.}
\label{tab:main_results}
\end{table*}

For RL optimization, we implement the GRPO-based training described in \S\ref{ssec:optimization} with the following default hyperparameters: learning rate $1 \times 10^{-6}$, global batch size $256$, $2$ epochs, $5$ rollouts per prompt, rollout temperature $1.0$, KL coefficient $\beta=0.001$, and clip ratio $\epsilon=0.2$.
For PCAR, we set the score-based modulation parameters $(\lambda_{\text{base}}, \lambda_{\text{max}})=(0.1,0.5)$ in Eq.~\ref{eq:pcar_lambda}, which determine the minimum and maximum strength of segment-wise rescaling.

\subsection{Main Results}
\label{ssec:main_results}

Table~\ref{tab:main_results} reports EM scores on seven open-domain QA benchmarks.
Across both backbone scales, \textsc{EvalAct} achieves the highest average EM among all compared methods, reaching 44.0\% with EvalAct-3B and 47.1\% with EvalAct-7B.
In both cases, it outperforms the second-best baseline, AutoRefine, by 3.5 and 1.6 points, respectively.

\paragraph{Comparison with Baselines.}
Compared with Search-o1, Search-R1, IRCoT, and Na\"ive RAG, \textsc{EvalAct} consistently outperforms these baselines on the majority of benchmarks, with the largest gains appearing in multi-hop settings.
This trend is consistent across both 3B and 7B backbones, indicating that the improvements are robust across model scales.
Unlike prior approaches that rely on implicit self-correction within free-form reasoning, \textsc{EvalAct} explicitly models evaluation as a discrete action, enabling segment-level credit assignment during RL optimization.

\paragraph{Multi-Hop Benchmarks.}
The strongest gains of \textsc{EvalAct} emerge on multi-hop datasets.
Across both backbone scales, \textsc{EvalAct} achieves the best performance on all four multi-hop benchmarks: 2WikiMultihopQA, Bamboogle, HotpotQA, and MuSiQue.
The gains are especially large on 2WikiMultihopQA and Bamboogle, where EvalAct-3B improves over AutoRefine by 10.6 and 13.6 points, respectively, and EvalAct-7B improves over the strongest baseline by 10.7 and 4.8 points.
Consistent improvements are also observed on HotpotQA and MuSiQue.
These results suggest that explicit intermediate evaluation is particularly beneficial for tasks requiring iterative evidence aggregation and long-horizon reasoning, where the coupled evaluation loop helps control error propagation across extended interaction sequences.

\paragraph{Single-Hop Benchmarks.}
On these single-hop datasets, \textsc{EvalAct} remains competitive but does not consistently outperform AutoRefine, which performs better on NQ and PopQA.
This is expected: AutoRefine is designed for iterative answer refinement, which is particularly effective when the main challenge is answer polishing rather than multi-step evidence accumulation in single-hop settings.
Nevertheless, the substantial gains on multi-hop benchmarks outweigh these gaps, resulting in the best overall average performance.

\section{Ablation Studies}
\label{ssec:ablation}

\begin{figure*}[htbp]
  \centering
  \includegraphics[width=0.98\linewidth]{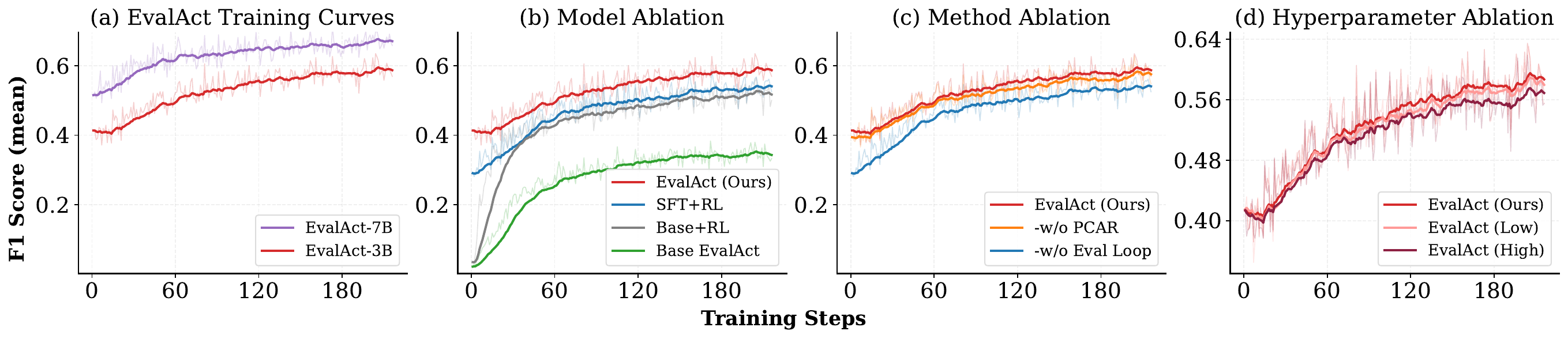}
  \caption{\textbf{Training curves and ablation overview.}
  (a) Training curves of \textsc{EvalAct} with 3B/7B backbones.
  (b) Model ablation across training variants.
  (c) Method ablation on removing the evaluation loop or PCAR.
  (d) Sensitivity to PCAR rescaling intensity.}
  \label{fig:ablation_overview}
\end{figure*}

We conduct ablation studies to understand the contribution of each component in \textsc{EvalAct}.
Figure~\ref{fig:ablation_overview} provides an overview: (a) training curves showing stable convergence for both 3B and 7B backbones, (b) model ablation comparing training variants, (c) method ablation isolating the evaluation loop and PCAR, and (d) sensitivity analysis of PCAR hyperparameters.
The following subsections detail these analyses.

\subsection{Model Ablation: Disentangling Format Alignment from Reasoning}
\label{sssec:ablation_model}

A prerequisite for \textsc{EvalAct} is protocol compliance: the agent must reliably produce well-formed tool calls and adhere to the strictly coupled \texttt{Search}$\rightarrow$\texttt{Evaluate} loop.
To disentangle the contributions of format acquisition (via SFT) from reasoning capability (via RL), we construct six training variants based on the Qwen2.5-3B-Instruct backbone and evaluate them on four multi-hop benchmarks.

The variants are defined as follows:
\begin{itemize}[leftmargin=*]
    \item \textbf{Base (Instruct) / SFT-Only:} the backbone model evaluated without and with supervised warm-up, respectively.
    \item \textbf{Base + RL / SFT + RL (Vanilla):} standard GRPO optimizing for answer correctness without enforcing the explicit \texttt{Evaluate} loop; retrieval tools are invoked freely.
    \item \textbf{Base EvalAct:} applied directly to the Base model without SFT warm-up.
    \item \textbf{EvalAct (Ours):} the full pipeline, i.e., SFT warm-up followed by \textsc{EvalAct} RL training.
\end{itemize}

\begin{figure}[htbp]
    \centering
    \includegraphics[width=\linewidth]{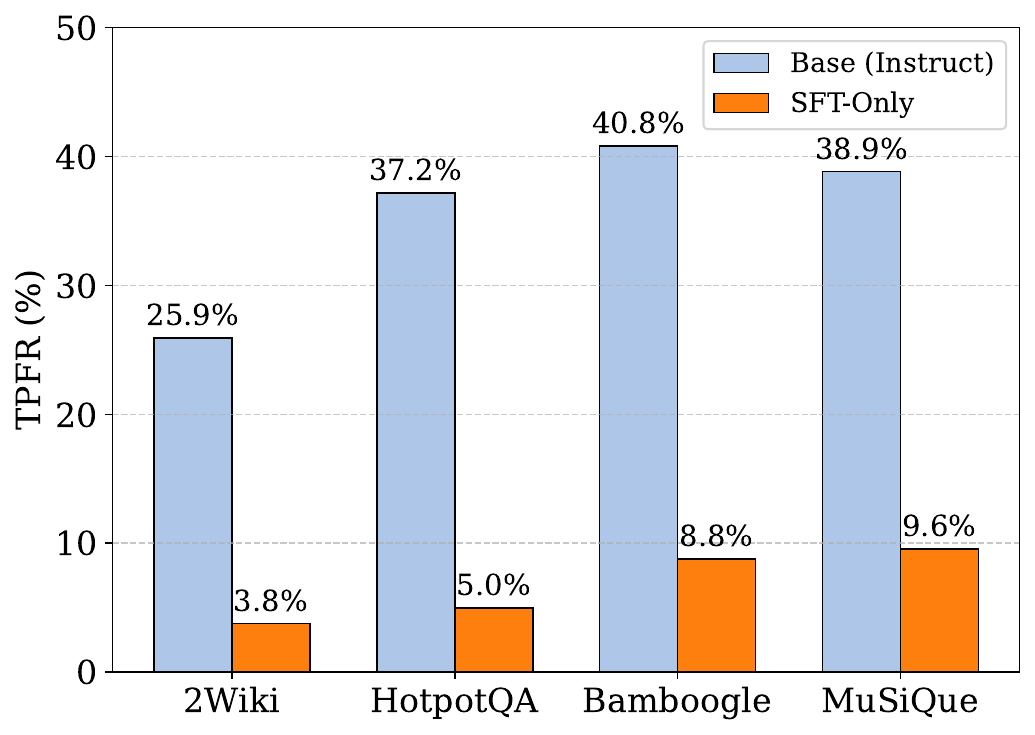}
    \caption{\textbf{Effect of SFT on Format Alignment.} The Base model exhibits high tool parsing failure rates.}
    \label{fig:tpfr_ablation}
\end{figure}

\begin{table}[htbp]
    \centering
    \resizebox{\columnwidth}{!}{%
    \begin{tabular}{l|cc|c}
        \toprule
        \textbf{Model Variant} & \textbf{SFT} & \textbf{RL Paradigm} & \textbf{Avg. EM} \\
        \midrule
        1. Base (Instruct) & \ding{55} & - & 14.2 \\
        2. SFT-Only & \ding{51} & - & 24.8 \\
        \midrule
        3. Base + RL (Vanilla) & \ding{55} & Standard & 33.1 \\
        4. SFT + RL (Vanilla) & \ding{51} & Standard & 33.5 \\
        \midrule
        5. Base EvalAct & \ding{55} & \textbf{EvalAct} & 17.1 \\
        6. \textbf{EvalAct (Ours)} & \ding{51} & \textbf{EvalAct} & \textbf{41.0} \\
        \bottomrule
    \end{tabular}
    }
    \caption{Ablation of training stages and paradigms on multi-hop benchmarks (Avg. EM, \%). \textit{Vanilla} denotes standard RL without the explicit evaluation loop.}
    \label{tab:ablation_main}
\end{table}

\paragraph{SFT for Format Alignment.}
We first examine the role of supervised warm-up in establishing protocol compliance.
As shown in Figure~\ref{fig:tpfr_ablation}, SFT substantially reduces tool-formatting and parsing failures, providing a stable initialization for structured tool use.
Consistent with this observation, SFT alone improves the multi-hop average from $14.2\%$ to $24.8\%$ (Table~\ref{tab:ablation_main}).
Under vanilla RL without the explicit evaluation loop, performance is only weakly affected by SFT initialization, reaching $33.1\%$ from the Base model and $33.5\%$ from the SFT-initialized model.
This suggests that standard RL can eventually recover a functional tool-calling policy, whereas SFT primarily stabilizes early optimization by aligning the model with the required format.

\paragraph{Effectiveness of the EvalAct Paradigm.}
We next isolate the effect of enforcing the coupled \texttt{Search}$\rightarrow$\texttt{Evaluate} loop.
With the same backbone and training budget, \textsc{EvalAct (Ours)} attains $41.0\%$ average EM, exceeding the strongest vanilla baseline ($33.5\%$) by $+7.5$ points (Table~\ref{tab:ablation_main}).
This gain supports the hypothesis that converting intermediate evaluation into an explicit action yields more informative process signals than implicit verification under outcome-only optimization.
By contrast, applying \textsc{EvalAct} directly without supervised warm-up yields only $17.1\%$, highlighting the difficulty of learning a structured action protocol from scratch.

\subsection{Method Ablation: Dissecting Structural and Optimization Components}
\label{sssec:ablation_method}

We further decompose the performance gains of \textsc{EvalAct} into two sources: the structural contribution of the explicit evaluation loop and the optimization contribution of Process-Calibrated Advantage Rescaling (PCAR).
We compare the full model against two ablated variants:
(1) \textbf{w/o Eval Loop:} removing \texttt{Evaluate} and reverting to a standard retrieval policy optimized via vanilla GRPO (equivalent to \textit{SFT+RL} in \S\ref{sssec:ablation_model});
(2) \textbf{w/o PCAR:} retaining the \texttt{Evaluate} structure and confidence scores $z$, but optimizing with standard GRPO without confidence-based advantage rescaling.

\begin{table}[htbp]
    \centering
    \resizebox{\columnwidth}{!}{%
    \begin{tabular}{l|cccc|c}
        \toprule
        \textbf{Method} & \textbf{2Wiki} & \textbf{Bam.} & \textbf{Hotpot} & \textbf{Mus.} & \textbf{Avg.} \\
        \midrule
        \textbf{EvalAct} & \textbf{50.0} & \textbf{48.0} & \textbf{44.3} & \textbf{21.6} & \textbf{41.0} \\
        \ w/o PCAR & 48.2 & 46.2 & 44.1 & 20.8 & 39.8 \\
        \ w/o Eval Loop & 41.4 & 37.2 & 38.9 & 16.5 & 33.5 \\
        \bottomrule
    \end{tabular}
    }
    \caption{Method ablation on multi-hop benchmarks (EM, \%).}
    \label{tab:ablation_method}
\end{table}

\paragraph{Structural Contribution of the Evaluation Loop.}
Removing the explicit evaluation mechanism causes the largest performance degradation.
As shown in Table~\ref{tab:ablation_method}, eliminating \texttt{Evaluate} lowers average EM from $41.0\%$ to $33.5\%$, a drop of $7.5$ points.
This degradation is consistent across all four benchmarks, with especially pronounced declines on 2WikiMultihopQA ($-8.6$) and Bamboogle ($-10.8$).
These results suggest that the primary benefit of \textsc{EvalAct} lies in its explicit evaluation loop, which enforces intermediate verification and thereby reduces error propagation in multi-hop reasoning.

\paragraph{Optimization Contribution of PCAR.}
Beyond the structural benefit of the explicit evaluation loop, PCAR provides additional optimization gains.
Compared with standard GRPO applied to the same evaluation-augmented framework, PCAR raises average EM from $39.8\%$ to $41.0\%$, a gain of $1.2$ points.
Improvements are observed on all four benchmarks, with gains of $1.8$ points on both 2WikiMultihopQA and Bamboogle, and smaller but consistent improvements on HotpotQA and MuSiQue.
These results indicate that confidence-aware advantage rescaling complements the explicit evaluation structure by providing more informative gradient signals for segments with varying reliability estimates.

\subsection{Hyperparameter Sensitivity}
\label{sssec:ablation_param}

We analyze the sensitivity of PCAR to the rescaling intensity governed by $\lambda_{\text{base}}$ and $\lambda_{\text{max}}$.
To characterize the strength of reliability-aware modulation, we define the \textbf{Relative Importance Ratio (RIR)} as the ratio between the maximum and minimum attainable advantage multipliers under full-confidence conditions.
When the standardized reliability score satisfies $\tilde{z}\in[-1,1]$, the unclamped multiplier spans $1\pm\lambda_{\text{max}}$; after applying the lower-bound clamp with threshold $\delta$, the effective ratio is approximated by $\text{RIR}\approx (1+\lambda_{\text{max}})\big/\max(\delta,\,1-\lambda_{\text{max}})$.

\begin{table}[htbp]
    \centering
    \resizebox{\columnwidth}{!}{%
    \begin{tabular}{l|cc|c|c}
        \toprule
        \textbf{Setting} & \textbf{Params} & \textbf{RIR} & \textbf{Avg.} & \textbf{$\Delta$} \\
        \midrule
        EvalAct (Low) & $0.05 / 0.25$ & $1.67$ & 39.8 & -0.3 \\
        \textbf{EvalAct (Mid)} & $\mathbf{0.10 / 0.50}$ & $\mathbf{3.0}$ & \textbf{40.1} & - \\
        EvalAct (High) & $0.20 / 1.00$ & $200$ & 39.4 & -0.7 \\
        \bottomrule
    \end{tabular}
    }
    \caption{Parameter ablation on PCAR intensity. \textbf{RIR} is the ratio between max/min multipliers.}
    \label{tab:ablation_param}
\end{table}

Table~\ref{tab:ablation_param} reports three representative configurations corresponding to low, moderate, and high rescaling intensity.
Performance remains relatively stable across settings, with average EM ranging from $39.4\%$ to $40.1\%$.
The moderate configuration (\textit{EvalAct (Mid)}, $\text{RIR}=3.0$) achieves the best overall performance.

Under conservative rescaling (\textit{Low}, $\text{RIR}=1.67$), the separation between high- and low-reliability segments is limited, potentially reducing the effectiveness of segment-level credit assignment.
Conversely, aggressive rescaling (\textit{High}, $\text{RIR}=200$) slightly degrades performance.
In this regime, the minimum multiplier approaches zero and is clipped by the lower-bound constraint, leading to highly imbalanced gradient magnitudes across segments.
Such extreme modulation can restrict corrective updates on low-reliability steps and destabilize optimization.

Overall, these results suggest that moderate reliability rescaling provides a balanced trade-off between emphasizing high-confidence segments and preserving sufficient gradient flow for error correction.

\section{Related Work}
\label{sec:related}

\subsection{Retrieval-Augmented Language Models}
Retrieval-augmented generation (RAG) enhances LLMs by grounding generation in externally retrieved knowledge~\citep{lewis2020retrieval,guu2020retrieval,borgeaud2022improving}.
Early work focused on improving retrieval quality through dense encoders~\citep{karpukhin2020dense,izacard2021leveraging} or neural rerankers~\citep{nogueira2019passage}.
More recent approaches integrate retrieval directly into the reasoning process, enabling models to iteratively query external sources~\citep{react,ircot,jiang2023active}.
Self-RAG~\citep{selfrag} introduces special tokens for assessing retrieval utility, though these remain implicit signals rather than structured actions.
Our work builds on this line by converting retrieval evaluation into an explicit, structured action with discrete scores that can serve as training signals.

\subsection{Reinforcement Learning for Tool-Using Agents}
RL has emerged as a promising approach for training LLM agents to use tools effectively~\citep{toolformer,nakano2021webgpt,qin2023toolllm}.
Search-R1~\citep{searchr1} demonstrates that pure RL with outcome rewards can train effective retrieval policies without supervised fine-tuning.
However, outcome-only rewards suffer from the credit assignment problem in multi-step trajectories.
Process reward models (PRMs) address this by providing step-level supervision~\citep{lightman2023let,uesato2022solving,wang2024math}, but typically require expensive human annotations or external verifiers that may not align with the target policy.
LeTS~\citep{zhang2025lets} designs retrieval-specific process rewards based on knowledge redundancy and exact match, yet relies on heuristics that may not generalize.

Recent work has begun to expand the action space by converting traditionally implicit behaviors into explicit, learnable decisions.
MemAct~\citep{zhang2025memory} formulates working memory management as policy actions for context deletion and insertion, enabling end-to-end RL over long-horizon context curation.
\textsc{EvalAct} shares this action-centric perspective but targets a different behavior: instead of treating retrieval quality assessment as an implicit part of free-form reasoning, we convert evaluation into an explicit action that produces process signals for fine-grained credit assignment.

\subsection{Self-Evaluation and Calibration}
LLMs can evaluate their own outputs~\citep{kadavath2022language,xie2023self,madaan2023self}, but such assessments vary in calibration—the alignment between expressed confidence and actual accuracy~\citep{tian2023fine}.
Uncalibrated evaluation during RL training risks reward hacking, where models assign high scores regardless of output quality.
Prior work addresses calibration through specialized training objectives~\citep{lin2022teaching}, prompting strategies~\citep{xiong2023can}, or post-hoc adjustments~\citep{zhao2021calibrate}.
We take a different approach: Process-Calibrated Advantage Rescaling (PCAR) designs the RL objective such that miscalibrated confidence incurs penalties through the advantage signal, naturally incentivizing well-calibrated evaluations without explicit calibration training.

\section{Conclusion}
\label{sec:conclusion}

We presented \textsc{EvalAct}, a framework that elevates retrieval evaluation from an implicit reasoning behavior to an explicit policy action. This design enables retrieval-augmented agents to generate structured process signals during interaction and to use them for more fine-grained reinforcement learning. Built on this framework, PCAR further improves optimization by aligning policy updates with segment-level reliability estimates. Across seven open-domain QA benchmarks, \textsc{EvalAct} delivers the best average results and shows its largest advantages on multi-hop reasoning tasks. These findings highlight the value of converting intermediate evaluation into a trainable action for multi-step retrieval-augmented reasoning.

\section*{Limitations}

This design directly addresses the two limitations identified above. 
At inference time, the evaluation output provides actionable control signals that facilitate early pruning of unproductive branches, reducing error propagation without external oracle supervision. 
During training, it produces dense, trajectory-aligned process signals that make intermediate reliability directly optimizable and enable finer-grained credit assignment.

\raggedbottom
\bibliography{custom}

\appendix

\section{Pseudocode for \textsc{EvalAct} with PCAR}
\label{sec:appendix_evalact_pcar}

\begin{algorithm}[H]
\caption{\textsc{EvalAct} Training with PCAR (GRPO Backbone)}
\label{app:EvalAct-PCAR}
\begin{algorithmic}[1]
\small
\STATE \textbf{Input:} policy $\pi_\theta$, reference $\pi_{\theta_{\text{ref}}}$, dataset $\mathcal{D}$, environment $\mathcal{E}$, \\
\hspace{2.2em} rollouts per input $G$, clip $\epsilon$, KL weight $\beta$, constant $\epsilon_0$, \\
\hspace{2.2em} PCAR params $(\lambda_{\text{base}}, \lambda_{\text{max}}, \delta)$
\STATE \textbf{Output:} optimized policy $\pi_\theta$

\WHILE{not converged}
    \STATE Sample a batch of queries $\mathcal{X} \sim \mathcal{D}$
    \STATE Initialize training buffer $\mathcal{B}\leftarrow\emptyset$

    \FORALL{$x \in \mathcal{X}$}
        \STATE // \textit{Protocol-compliant rollouts}
        \FOR{$i=1$ \TO $G$}
            \STATE Sample trajectory $y_i \sim \pi_{\theta_{\text{old}}}(\cdot\mid x)$ under coupled $\texttt{Search}\!\rightarrow\!\texttt{Evaluate}$
            \STATE Compute gated reward $r_i \leftarrow \mathcal{G}(y_i,y^*)$ (Eq.~\eqref{eq:reward})
            \STATE Segment $y_i$ into $\{\sigma_{i,k}\}_{k=1}^{K_i}$ aligned with $\texttt{Search}\!\rightarrow\!\texttt{Evaluate}$, record $\{z_{i,k}\}_{k=1}^{K_i}$
        \ENDFOR

        \STATE // \textit{GRPO advantage}
        \STATE $\mu_{\text{group}}\leftarrow \mathrm{mean}(\{r_i\}_{i=1}^{G})$, \quad
               $\sigma_{\text{group}}\leftarrow \mathrm{std}(\{r_i\}_{i=1}^{G})+\epsilon_0$
        \FOR{$i=1$ \TO $G$}
            \STATE $A_i \leftarrow (r_i-\mu_{\text{group}})/\sigma_{\text{group}}$
            \STATE Compute calibrated advantages $\{\hat{A}_{i,t}\}$ for tokens in $y_i$ via \textsc{PCAR}:
            \STATE \hspace{1.4em} $\tilde z_{i,k} \leftarrow (z_{i,k}-\mu_i)/(\sigma_i+\epsilon_0)$,\;
                  $\lambda_{i,k}\leftarrow \lambda_{\text{base}}+(\lambda_{\text{max}}-\lambda_{\text{base}})\cdot z_{i,k}/10$
            \STATE \hspace{1.4em} $\hat{A}_{i,t} \leftarrow A_i \cdot \mathrm{clamp}(1+\lambda_{i,k}\tilde z_{i,k},\delta,\infty)$ for $t\in\sigma_{i,k}$
            \STATE Add token-level instances from $(x,y_i)$ with advantages $\{\hat{A}_{i,t}\}$ to $\mathcal{B}$
        \ENDFOR
    \ENDFOR

    \STATE Update $\pi_\theta$ by maximizing the clipped objective with KL regularization (Eq.~\eqref{eq:pcar_objective}) on $\mathcal{B}$
\ENDWHILE

\STATE \textbf{return} $\pi_\theta$
\end{algorithmic}
\end{algorithm}

\section{Evaluate Specification and Feedback Templates}
\label{app:evaluate_templates}

\paragraph{\texttt{Evaluate} Action Format.}
After each retrieval step, the agent invokes $\texttt{Evaluate}(c,z)$, where $c$ is a free-form textual assessment of the immediately preceding \texttt{Search} output, and $z\in[0,10]$ denotes a scalar self-reported confidence score.
The environment intercepts this action and returns a discrete control cue $\mathcal{I}=\Phi(z)$, which is appended to the context to modulate subsequent decision-making.

\paragraph{Discrete Feedback Mapping.}
We instantiate $\Phi(\cdot)$ as a deterministic, three-tier binning strategy:
\begin{equation}
\Phi(z)=
\begin{cases}
\mathcal{I}_{\text{low}}, & z \in [0,3],\\[2pt]
\mathcal{I}_{\text{mid}}, & z \in (3,7],\\[2pt]
\mathcal{I}_{\text{high}}, & z \in (7,10].
\end{cases}
\end{equation}

\paragraph{Feedback Templates.}
The control cue $\mathcal{I}$ is materialized as an instruction-style message that conditions the agent's next action. The exact textual templates returned by the environment are detailed below.

\begin{tcolorbox}[breakable, title={\textbf{Environment Feedback Templates}}, colframe=darkgray, colback=gray!5, top=2mm, bottom=2mm, left=3mm, right=3mm, fontupper=\small]

\textbf{$\mathcal{I}_{\text{low}}$ (Low Quality $\mid$ $z \in [0,3]$)} \\
\textit{``The previous \texttt{Search} results are largely irrelevant or unhelpful for answering the question. Do not rely on them. Reformulate the query (e.g., alternative keywords or a different angle) and issue a new \texttt{Search}.''}

\noindent\textcolor{gray!50}{\rule{\linewidth}{0.4pt}}

\textbf{$\mathcal{I}_{\text{mid}}$ (Medium Quality $\mid$ $z \in (3,7]$)} \\
\textit{``The previous \texttt{Search} results contain partially useful evidence but may be incomplete or noisy. Use only clearly relevant excerpts. Consider an additional, more targeted \texttt{Search} to fill missing details, resolve remaining subproblems, or verify uncertain information.''}

\noindent\textcolor{gray!50}{\rule{\linewidth}{0.4pt}}

\textbf{$\mathcal{I}_{\text{high}}$ (High Quality $\mid$ $z \in (7,10]$)} \\
\textit{``The previous \texttt{Search} results are highly relevant and constitute substantive progress toward answering the question (e.g., providing key facts or resolving an important subtask). Use them as primary evidence to construct the answer. Perform another \texttt{Search} only if a specific critical detail is still missing.''}

\end{tcolorbox}

\section{Case Study: Multi-Hop Reasoning Trajectory}
\label{app:case_study}

We present a full multi-hop interaction trajectory to illustrate the \textsc{EvalAct} framework in practice. The example highlights calibrated self-evaluation: a partial retrieval receives 5/10, while a conclusive retrieval receives 10/10. These step-wise confidence signals are the same signals used by PCAR for segment-level advantage rescaling during RL.

\begin{tcolorbox}[breakable, title={\textbf{EvalAct Trajectory Example}}, colframe=darkgray, colback=gray!5, top=2mm, bottom=2mm, left=3mm, right=3mm, fontupper=\small]

\textbf{User:} In between \textit{Remember the Titans} and \textit{My Favorite Martian} which film grossed \$36.8 million domestically?

\vspace{0.5em}
\textcolor{blue!80!black}{\textbf{Agent:}} \\
\texttt{<think>}I need to find out which film grossed \$36.8 million domestically between "Remember the Titans" and "My Favorite Martian". I should search for box office information for both films.\texttt{</think>}

\vspace{0.5em}
\textcolor{teal!80!black}{\textbf{Tool Call [Search]:}} \\
\texttt{\{"query": "Remember the Titans domestic box office \$36.8 million"\}}

\vspace{0.5em}
\textcolor{gray!80!black}{\textbf{Observation [Search]:}} \\
\textit{[Doc 1 \& 3 omitted for brevity]} \\
Doc 2 (Title: "Remember the Titans"): "...Remember the Titans opened strongly at the U.S. box office... It eventually went on to gross an estimated \$115,654,751 in the U.S., and..."

\vspace{0.5em}
\textcolor{teal!80!black}{\textbf{Tool Call [Evaluate]:}} \\
\texttt{\{} \\
\hspace*{1em}\texttt{"evaluation": "The search provided information about Remember the Titans' box office performance, but didn't specifically mention the \$36.8 million figure. The results show it grossed \$115,654,751 in the U.S., which is much higher than \$36.8 million. This helps narrow down that Remember the Titans didn't gross \$36.8 million, but I need more specific information to compare with My Favorite Martian.",} \\
\hspace*{1em}\texttt{\textbf{"score": 5}} \\
\texttt{\}}

\vspace{0.5em}
\textcolor{gray!80!black}{\textbf{Observation [Evaluate]:}} \\
\textbf{Score 5/10 (Medium Quality):} The previous Search results are partially helpful but may be incomplete or contain noise. Extract and use only the clearly relevant parts. Consider running another more focused Search to fill in missing details, resolve remaining subproblems, or verify uncertain information.

\hdashrule{\linewidth}{0.5pt}{1pt}

\textcolor{blue!80!black}{\textbf{Agent:}} \\
\texttt{<think>}Now I need to search for My Favorite Martian's box office information.\texttt{</think>}

\vspace{0.5em}
\textcolor{teal!80!black}{\textbf{Tool Call [Search]:}} \\
\texttt{\{"query": "\textbackslash"My Favorite Martian\textbackslash" domestic box office \$36.8 million"\}}

\vspace{0.5em}
\textcolor{gray!80!black}{\textbf{Observation [Search]:}} \\
\textit{[Doc 1 \& 2 omitted for brevity]} \\
Doc 3 (Title: "My Favorite Martian (film)"): "...The film grossed \$36.8 million domestically against a budget of \$65 million..."

\vspace{0.5em}
\textcolor{teal!80!black}{\textbf{Tool Call [Evaluate]:}} \\
\texttt{\{} \\
\hspace*{1em}\texttt{"evaluation": "This search was highly successful. Document 3 directly mentions \textbackslash"My Favorite Martian\textbackslash" and states it grossed \$36.8 million domestically against a budget of \$65 million. This matches exactly the figure mentioned in the question and confirms this film's domestic gross. Combined with the previous search showing Remember the Titans grossed much more than \$36.8 million (\$115+ million), this gives me clear information to answer which film grossed the specified amount.",} \\
\hspace*{1em}\texttt{\textbf{"score": 10}} \\
\texttt{\}}

\vspace{0.5em}
\textcolor{gray!80!black}{\textbf{Observation [Evaluate]:}} \\
\textbf{Score 10/10 (High Quality):} The previous Search results are highly relevant and make strong progress toward answering the question... Use them as primary evidence to construct your answer. Only conduct another Search if you still need a very specific missing detail.

\hdashrule{\linewidth}{0.5pt}{1pt}

\textcolor{blue!80!black}{\textbf{Agent:}} \\
\texttt{<think>}From the information gathered: \\
1. Remember the Titans grossed \$115,654,751 in the U.S. (much higher than \$36.8 million) \\
2. My Favorite Martian grossed \$36.8 million domestically \\
Therefore, My Favorite Martian is the film that grossed \$36.8 million domestically.\texttt{</think>}

\vspace{0.5em}
\texttt{<answer>}My Favorite Martian\texttt{</answer>}

\end{tcolorbox}

\end{document}